\documentclass[journal]{IEEEtran}
\usepackage[labelfont=bf]{caption}
\usepackage{graphicx}
\usepackage{amsmath}
\usepackage{amssymb}
\usepackage{breqn}
\usepackage{booktabs}
\usepackage{multirow}
\usepackage{subfloat}
\usepackage{pifont}
\newcommand{\cmark}{\ding{51}}%
\newcommand{\xmark}{\ding{55}}%
\usepackage[dvipsnames]{xcolor}
\usepackage{svg}
\usepackage{tikz}
\usepackage{comment}
\usepackage{color}
\usepackage[arrowdel]{physics}
\usepackage{pgfplots, pgfplotstable}
\usepackage{cite}
\usepackage[pagebackref,breaklinks,colorlinks]{hyperref}
\usepackage[accsupp]{axessibility}
\usepackage{floatrow}
\newfloatcommand{capbtabbox}{table}[][\FBwidth]
\usepackage{blindtext}
\usepackage{float}
\floatstyle{plaintop}
\restylefloat{table}

\usepackage[pagebackref,breaklinks,colorlinks]{hyperref}

\usepackage[capitalize]{cleveref}
\crefname{section}{Sec.}{Secs.}
\Crefname{section}{Section}{Sections}
\Crefname{table}{Table}{Tables}
\crefname{table}{Tab.}{Tabs.}

\begin{document}

\title{Deep Metric Learning for Unsupervised Remote Sensing Change Detection
}

\author{Wele Gedara Chaminda Bandara,~\IEEEmembership{Student Member,~IEEE,} and Vishal M. Patel,~\IEEEmembership{Senior Member,~IEEE}
\thanks{The authors are with the Department
of Electrical and Computer Engineering, Johns Hopkins University, Baltimore,
MD, 21218 USA e-mail: \{wbandar1, vpatel36\}@jhu.edu.}
}

\markboth{IEEE Transactions on Geoscience and Remote Sensing}%
{Shell \MakeLowercase{\textit{et al.}}: Bare Demo of IEEEtran.cls for Journals}

\maketitle

\begin{abstract}
Remote Sensing Change Detection (RS-CD) aims to detect relevant changes from Multi-Temporal Remote Sensing Images (MT-RSIs), which aids in various RS applications such as land cover, land use, human development analysis, and disaster response. The performance of existing RS-CD methods is attributed to training on large annotated datasets. Furthermore, most of these models are less transferable in the sense that the trained model often performs very poorly when there is a domain gap between training and test datasets. This paper proposes an unsupervised CD method based on deep metric learning that can deal with both of these issues. Given an MT-RSI, the proposed method generates corresponding change probability map by iteratively optimizing an unsupervised CD loss without training it on a large dataset. Our unsupervised CD method consists of two interconnected deep networks, namely Deep-Change Probability Generator (D-CPG) and Deep-Feature Extractor (D-FE). The D-CPG is designed to predict change and no change probability maps for a given MT-RSI, while D-FE is used to extract deep features of MT-RSI that will be further used in the proposed unsupervised CD loss. We use transfer learning capability to initialize the parameters of D-FE. We iteratively optimize the parameters of D-CPG and D-FE for a given MT-RSI by minimizing the proposed unsupervised ``\textit{similarity-dissimilarity loss}''.  This loss is motivated by the principle of \textit{metric learning} where we simultaneously maximize the distance between change pair-wise pixels while minimizing the distance between no-change pair-wise pixels in bi-temporal image domain and their deep feature domain. The experiments conducted on three CD datasets show that our unsupervised CD method achieves significant improvements over the state-of-the-art supervised and unsupervised CD methods. Code available at \href{https://github.com/wgcban/Metric-CD}{https://github.com/wgcban/Metric-CD}
\end{abstract}

\section{Introduction}
\begin{figure}[tb]
    \centering
    \includegraphics[width=\linewidth]{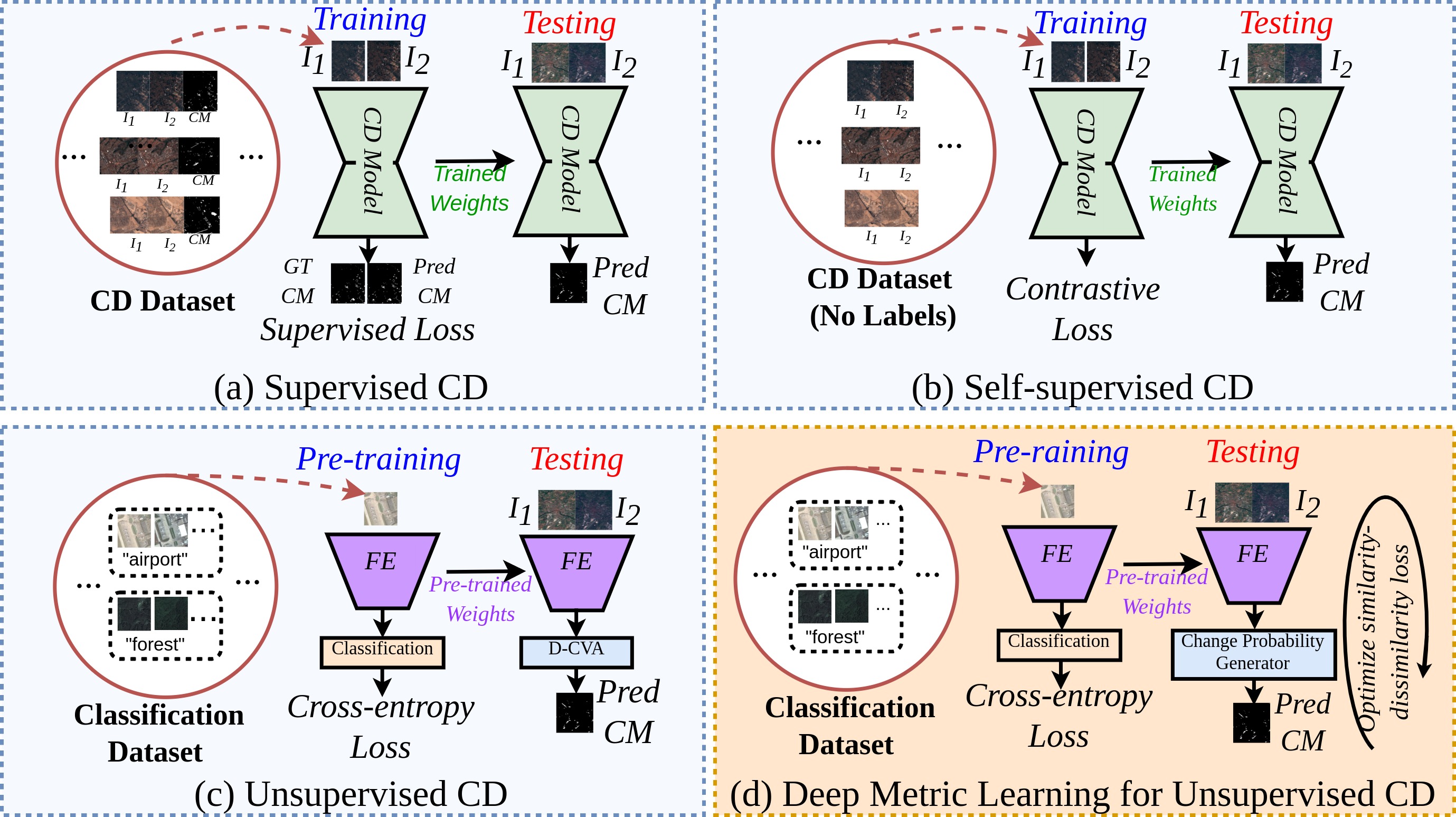}
    \caption{How our deep metric learning-based  CD method differs from the  state-of-the-art CD methods. \textbf{(a)  Supervised CD}: In this case, MT-RSIs and the corresponding change masks are available for training the CD model \cite{wiratama2019fusion,daudt2018fully,SeCo}. \textbf{(b) Self-supervised CD}: In this case, change masks are not available during training, but a large unlabeled MT-RSI dataset is utilized to synthesize changes \cite{chen2021self,ren2020unsupervised}. \textbf{(c) Unsupervised CD}: These methods start with a sub-optimal pre-trained Feature Extractor (FE) that is trained for some other task to obtain deep features, and then perform Change Vector Analysis (CVA) followed by thresholding to obtain change maps \cite{8608001,de2019unsupervised}. \textbf{(d) Deep  Metric Learning for Unsupervised CD (ours):} We iteratively optimize the parameters of deep change probability generator (D-CPG) to obtain change and no-change probability maps for a given MT-RSI according to the proposed unsupervised ``\textit{similarity-dissimilarity}'' loss for CD. The ``\textit{similarity-dissimilarity}'' loss is motivated by the popular principle of metric learning where we maximize the distance between change pair-wise pixels and minimize the distance between no-change pairwise pixels in bi-temporal image domain and deep feature domain.
    }
    \label{fig:intro}
\end{figure}

Change detection (CD) aims to detect \textit{relevant} changes from multi-temporal remote sensing images (MT-RSIs) captured over the same geographic area at distinct times.  In recent years, CD has gained great attention in the research community due to its numerous applications in urban planning, environmental monitoring, disaster assessment, agriculture, forestry, etc. \cite{singh1989review,coppin2004review}. Most of these practical applications typically need to quickly obtain the relevant change information from  large MT-RSIs. In the context of CD, the \textit {irrelevant} changes such as vegetation color variations, building shadows, atmospheric variations, lighting changes, and cloud covering limit the accuracy of change maps.

Most CD techniques proposed in the literature are based on \textit{supervised} training where a model (i.e., classical or deep learning based model) is trained with a large number of MT-RSIs and corresponding change labels (see Fig. \ref{fig:intro}-(a)). However, in MT problems, it is expensive and often impossible to obtain a large number of annotated training samples for modeling change and no change classes \cite{wiratama2019fusion,daudt2018fully,SeCo,9883686,diffusion_cd}. This limited access to labeled training data has motivated the researchers to explore self-supervised and unsupervised CD methods. 

The recently proposed \textit{self-supervised} methods are often based on either self-supervised pre-training \cite{chen2021self, diffusion_cd} or contrastive learning \cite{ren2020unsupervised} (see Fig. \ref{fig:intro}-(b)). In self-supervised pre-training a model is trained by utilizing the change labels extracted directly from the MT-RSIs themselves. In other words, instead of training a model for the CD task, self-supervised pre-training optimizes a model for a pretext task in which the labels are extracted directly from the MT-RSIs by alternating relative positions of patches. By doing that, the network is pre-trained to recognize the changes in MT-RSIs with no annotated change masks. In contrastive self-supervised CD, a pseudo-Siamese network is trained to regress the output between its two branches, which is pre-trained in a contrastive way on a large unlabeled MT-RSI dataset. However, both of these self-supervised CD methods require a large unlabeled MT-RSI dataset for optimizing the model parameters. In addition, the performance of the trained model heavily depends on how realistic the changes they make while training the model.

In contrast to self-supervised CD methods, \textit{unsupervised} CD methods \cite{8608001,de2019unsupervised} do not often depend on a large MT-RSI dataset for training.  They extract feature representations of a given MT-RSI from a Feature Extractor (FE) which is initialized from a network pre-trained for some other task (like classification, segmentation, etc.) via transfer learning (see Fig. \ref{fig:intro}-(c)). Next, Change Vector Analysis (CVA)~\cite{DeepCVA} or Deep Feature Difference (D-FD)~\cite{QuickBird} is performed on the extracted deep features to identify the change pixels. However, the existing deep unsupervised CD methods are sensitive to pseudo \textit{unwanted} changes, because the feature extractor is not trained to produce deep features which are invariant to color and seasonal transformations that usually appear in MT-RSIs. 

\begin{figure}[tb]
    \centering
    \includegraphics[width=\linewidth]{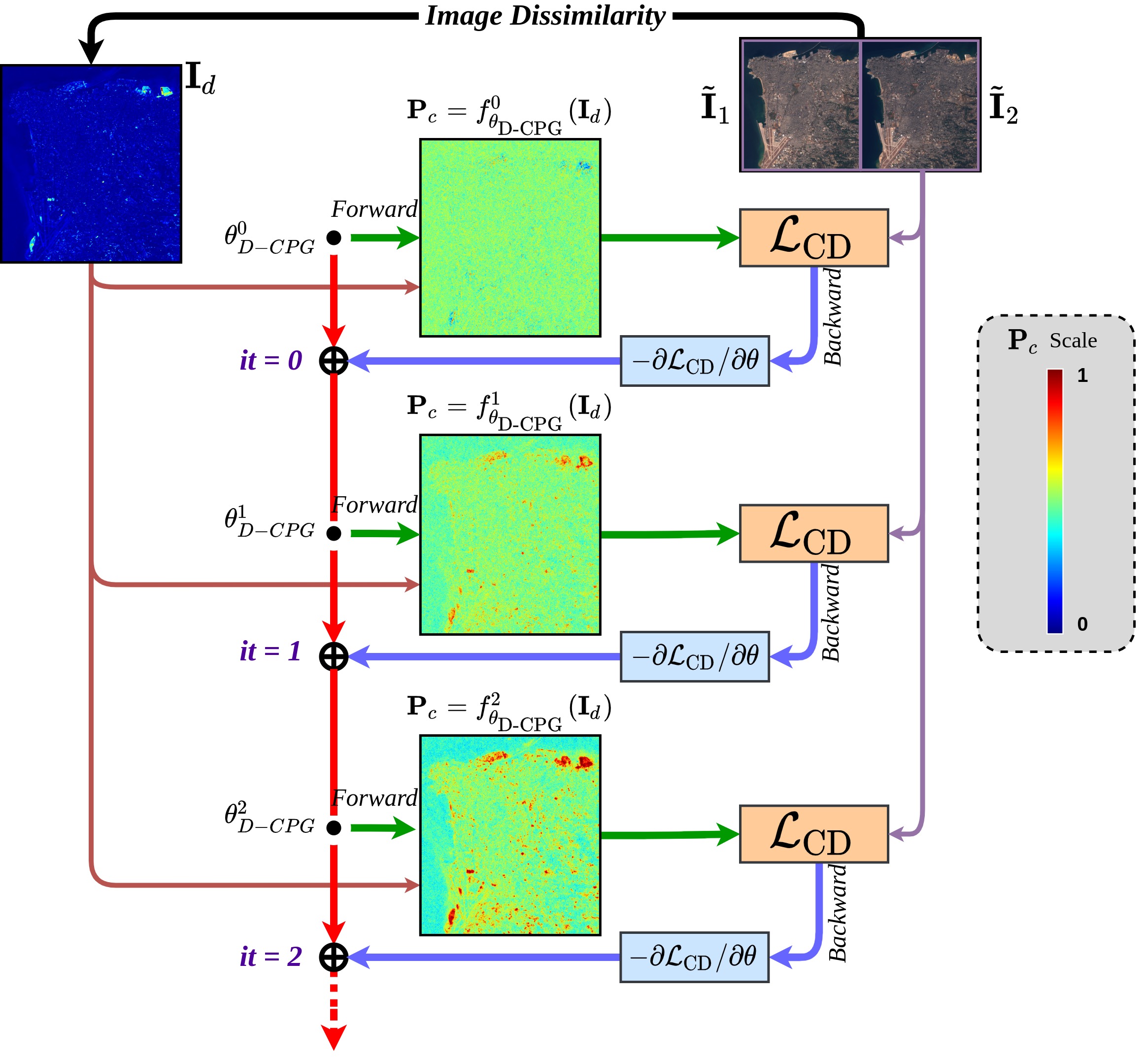}
    \caption{We start from random weights for D-CPG $\theta^0_{\text{D-CPG}}$, and iteratively update the weights in order to minimize the proposed $\mathcal{L}_{\text{CD}}$. At the $t$-th iteration, weights $\theta^t_{\text{D-CPG}}$ are used to generate the change probability maps $\mathbf{P}_{\text{c}}^t = f_{\theta_{\text{D-CPG}}}(\mathbf{I}_d)$.  $\mathbf{P}_{\text{c}}^t$ is used to calculate $\mathcal{L}_{\text{CD}}$. The gradient of the loss  $\mathcal{L}_{\text{CD}}$ with respect to the weights $\theta$ is then computed and used to update the parameters.} 
    \label{fig:itterative_opt}
\end{figure}

In order to address the aforementioned issues of state-of-the-art (SOTA) CD methods, we propose a novel unsupervised \textit{deep metric learning}-based strategy for CD (see Fig. \ref{fig:intro} - (d)). Our CD method consists of two networks, namely Deep-Change Probability Generator (D-CPG) and Deep-Feature Extractor (D-FE). The parameters of D-CPG are randomly initialized while the parameters of D-FE are initialized via transfer learning. The weights of both D-CPG and D-FE are iteratively optimized for a given MT-RSI according to the proposed \textit{similarity-dissimilarity loss} which is motivated by the principal of metric learning where we maximize the similarity between no-change pair-wise pixels while minimizing the similarity between change pair-wise pixels in bi-temporal image domain and their respective deep feature domain as shown in Fig. \ref{fig:itterative_opt}. Furthermore, to force the deep features of the given MT-RSI from the D-FE to be invariant to local colorimetric variations, we also define a self-supervised auxiliary task in which we force the feature representations of a given MT-RSI to be consistent under different random color transformations through the auxiliary \textit{context consistency loss}. In our experiments, we show the significance of the proposed CD method by comparing it with SOTA supervised, self-supervised, and unsupervised CD methods.

\section{Related Works}
\paragraph{Classical CD methods.} Although there are many classical methods available for CD from RSIs, most of them are based on the Difference Image (DI). A DI can be obtained by image differentiation \cite{st2014subsense}, image ratio \cite{skifstad1989illumination}, Principal Component Analysis (PCA) \cite{kuncheva2013pca}, or Change Vector Analysis (CVA) \cite{lu2004change}. Once the DI is calculated, the binary change map is obtained by classifying the DI into two classes, namely change class and no change class. This can be achieved by thresholding the DI with an appropriate threshold \cite{st2014subsense}, using a clustering algorithm (i.e., K-Means \cite{celik2009unsupervised,hartigan1979algorithm}, metric learning \cite{hartigan1979algorithm}, etc.), or a supervised classification \cite{walter2004object}. However, the main drawback of these classical approaches is that the resulting change map contains many pseudo-random changes due to isolated radiometric changes, vegetation color variations, and color changes due to seasonal effects \cite{8937747}. Although supervised classification is the best way to avoid such random changes,  it requires expertly annotated pixel-level labels, limiting the applicability of many conventional approaches.

\paragraph{Deep Learning-based CD methods.} Recently, a variety of methods based on Convolutional Neural Networks (ConvNets) have also been proposed for CD from MT-RSIs due to their excellent ability to learn low-level and high-level features automatically during training \cite{shi2020change,khelifi2020deep,semicd}. Among these, most of them are fully supervised methods.  They often require more training samples and more training time to learn the deep features of RSIs \cite{di2021deep}. Early Fusion (EF) \cite{wiratama2019fusion}, and  Siamese networks \cite{daudt2018fully} are the widely adopted architectures for supervised CD. However, collecting a large number of training samples that require learning deep features is difficult to achieve in many CD applications, especially with high-precision and pixel-level annotations \cite{khelifi2020deep}. In recent years, the limited access to labeled change information has increased the researcher's attention to self-supervised and unsupervised CD. Self-supervised pre-training (SSP-CD) \cite{leenstra2021self}, Generative Adversarial Networks \cite{ren2020unsupervised} (GAN-CD),  and contrastive loss-based pseudo-Siamese networks (CPS-CD) \cite{chen2021self}  are the widely used self-supervised CD methods. Even though self-supervised learning encourages the network to learn meaningful features, it requires a large unlabeled dataset that may not be available for some CD applications. Most unsupervised CD methods are based on transfer learning \cite{pan2009survey}, where they extract deep features of MT-RSIs by initializing network weights that are pre-trained for some other task (i.e., segmentation or classification) on real or RS images. Examples of unsupervised-CD include Deep - Feature Difference (FeatureDiff) \cite{8608001} and Deep - Change Vector Analysis (D-CVA) \cite{de2019unsupervised}. Once the deep features are obtained, FeatureDiff and D-CVA obtain the change map by computing the feature difference or performing CVA followed by thresholding, respectively. 

\section{Proposed Method}
\label{sec: method}
\subsection{Problem formulation}
\par Consider two co-registered \textit{multi-spectral} remote sensing images $\mathbf{I}_1$ and $\mathbf{I}_2$ with $b$ bands, captured over the same geographical area at distinct time instances $t_1$ and $t_2$, respectively. Usually $\mathbf{I}_1$ and $\mathbf{I}_2$ are referred to as the pre-change image and post-change image, respectively and $\{\mathbf{I}_1, \mathbf{I}_2\}$ is referred to as the multi-temporal image. Our objective is to detect the changes from $\mathbf{I}_1$ and $\mathbf{I}_2$ in an unsupervised way; i.e., we classify the set of all pixels ($\bf \Omega$) into change ($\mathbf{\Omega}_{\textit{c}}$) and no-change ($\mathbf{\Omega}_{\textit{nc}}$) pixels. If we compute the dissimilarity between each pixel of $\mathbf{I}_1$ and $\mathbf{I}_2$ individually, we often end up obtaining either not-meaningful changes from a \textit{context-based perspective} or changes which are minute to be considered as a change of interest because of inherent radiometric dissimilarities between pre-change and post-change images. Such small isolated radiometric changes are therefore not considered as a change.

\subsection{Overview of the proposed solution}
\par
\begin{figure*}[tb]
    \centering
    \includegraphics[clip, trim=0 0 1cm 0, width=\linewidth]{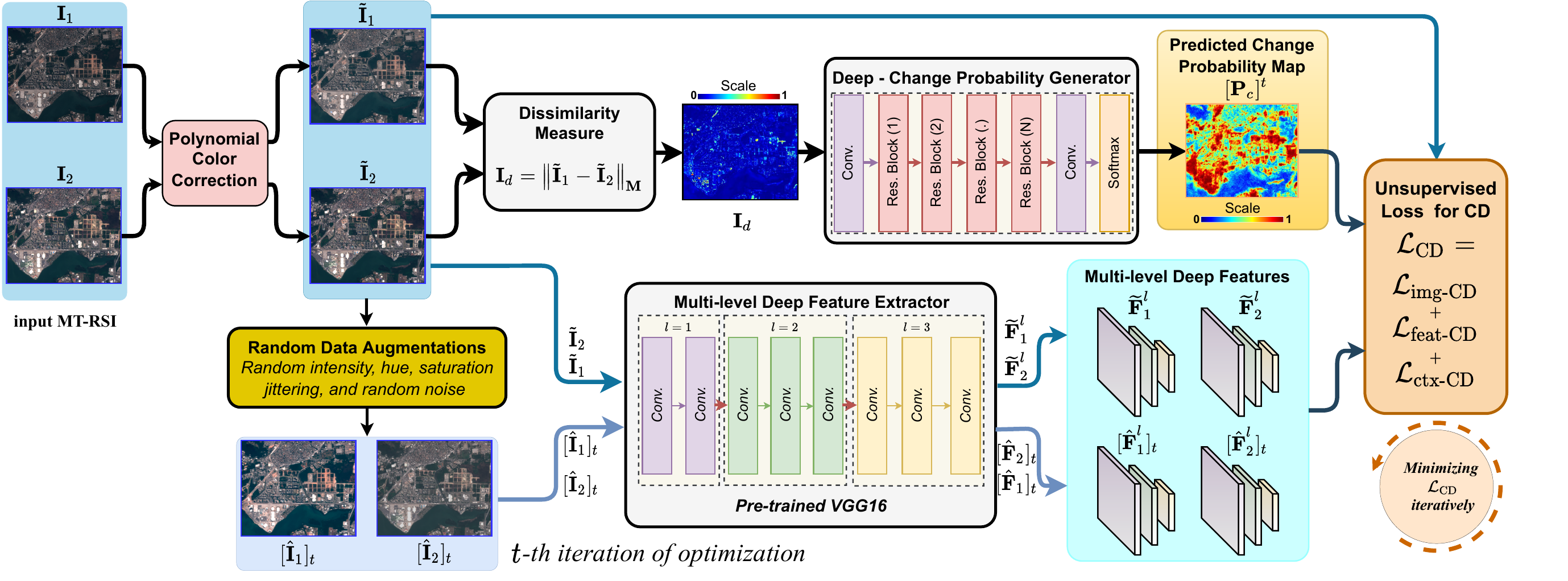}
    \caption{The proposed unsupervised deep metric learning framework for CD in multi-temporal remote sensing images.
    }
    \label{fig:method}
\end{figure*}

 The proposed unsupervised framework addresses the aforementioned issues of change detection. As depicted in Fig. \ref{fig:method}, we first perform polynomial color correction on multi-temporal image $\{\mathbf{I}_1, \mathbf{I}_2\}$ to obtain pre-processed multi-temporal image $\{\widetilde{\mathbf{I}}_1, \widetilde{\mathbf{I}}_2\}$. Next, we compute the difference image $\mathbf{I}_{d}$  and process it through a Deep - Change Probability Generator (D-CPG) to generate the change probability map $\mathbf{P}_{c}$. The weights of D-CPG are optimized by minimizing the proposed \textit{similarity-dissimilarity} loss in an unsupervised manner. However, minimizing the similarity-dissimilarity loss in image-domain results in detecting unwanted changes due to local radiometric shifts of pre-change and post-change images. To make our change detection framework robust to such isolated radiometric changes, we also utilize similarity-dissimilarity loss on multi-level feature representations of $\{\widetilde{\mathbf{I}}_1, \widetilde{\mathbf{I}}_2\}$, obtained from a Deep Feature Extractor (D-FE) that is pre-trained for some other task. In order to constrain features from D-FE to be robust to isolated color shifts, we apply data augmentations on $\{\widetilde{\mathbf{I}}_1, \widetilde{\mathbf{I}}_2\}$ and constrain their deep-features to be consistent with deep-features of  $\{\widetilde{\mathbf{I}}_1, \widetilde{\mathbf{I}}_2\}$ via the context-consistency loss. We iteratively minimize the overall loss function $\mathcal{L}_{\text{CD}}$ for unsupervised CD to generate optimal change probability map $\mathbf{P}^*_c$ for a given multi-temporal image $\{\mathbf{I}_1, \mathbf{I}_2\}$ as shown in fig. \ref{fig:itterative_opt}.

\subsection{Proposed unsupervised CD framework}

\paragraph{Image preprocessing and colorimetric mapping.}
We adopt Polynomial Color Correction (PCC)\cite{hong2001study} to reduce colorimetric changes due to different sensor properties, seasonal variations, and lighting conditions. It has been experimentally shown that PCC can significantly reduce colorimetric errors \cite{deep_wb,root_pcc}. Using PCC, we map the pre-change image color space to the post-change image color space using a polynomial mapping function $\bf M$, usually found by Least Squares Regression (LSR)\cite {root_pcc}. To do that, we first take $ N_c $ color samples (pixels) from each pre-change and post-change image by downsampling the original images by a factor of $ \beta $. We denote the downsampled versions of pre-change and post-change images as $ \mathbf{i}_1 $ and $ \mathbf{i}_2 $, respectively. The reason for applying LSR on downsampled image pixels is to more accurately bring out the color properties of the images than the actual changes present in multi-temporal image. Also, this will significantly reduce the pre-processing time especially when the satellite images have a very high spatial resolution. Consequently, we compute the polynomial mapping matrix $\bf M$ that maps values of $\rho (\mathbf{i}_1) $ to the colors of the post-change image $ \mathbf{i}_2 $, where $ \rho (\cdot) $ is a polynomial kernel function that maps the spectral vectors of the satellite image to a higher $d$-dimensional space. See the supplementary document for an assessment of the different kernel functions. The mapping matrix $ \bf M $ is found by least squares regression and has a closed form solution of the form:
\begin{equation}
    \mathbf{M} =  (\rho(\mathbf{i}_1)^T\rho(\mathbf{i}_1))^{-1} \rho(\mathbf{i}_1)^T \mathbf{i}_2.
\end{equation}
Once $\bf M$ is computed, we obtain our final color transformed version of pre-change and post-change images as follows:
\begin{equation}
    \widetilde{\mathbf{I}}_1 = \mathbf{M} \rho(\mathbf{I}_1)\text{ , and } \widetilde{\mathbf{I}}_2 = \mathbf{I}_2.
\end{equation}

\paragraph{Difference Image $\mathbf{I}_d$.} 
We compute the dissimilarity measure $\mathbf{I}_d$ between pre-change $(\mathbf{\widetilde{I}}_1)$ and post-change $(\mathbf{\widetilde{I}}_2)$ images by utilizing the Mahalanobis distance \cite{de2000mahalanobis} as follows:
\begin{equation}
    \mathbf{I}_d = \left\| \widetilde{\mathbf{I}}_1 - \widetilde{\mathbf{I}}_2\right\|_{\mathbf{M}} = \sqrt{\left(\widetilde{\mathbf{I}}_1-\widetilde{\mathbf{I}}_2\right)^T \mathbf{S}^{-1}  \left(\widetilde{\mathbf{I}}_2-\widetilde{\mathbf{I}}_2\right)},
\end{equation}
where $\bf S$ is the covariance matrix of size $b \times b$, and $T$ denotes the matrix transpose. We use the  Mahalanobis distance since it computes the dissimilarity measure based on the distribution pattern of data points through the covariance matrix $\mathbf{S}$, unlike $L_1$ or $L_2$ distances. The conventional CD methods use the difference image $\mathbf{I}_d$ and threshold it with an appropriate generic rule to obtain the binary change map \cite{radke2005image}. However, the latter case is sensitive to noise and local illumination variations, and does not consider local consistency properties and object-based perspective of the change
mask.

\paragraph{Deep-Change Probability Generator (D-CPG).} 
To enforce the local consistency property of the change mask, we go one step beyond the conventional CD methods where we take the difference image $\mathbf{I}_d$ as input and process it through a Deep - Change Probability Generator (D-CPG) network. The D-CPG estimates the probability of change for a given pixel location $ (i, j) - \mathbf {P} _c (i, j)$ by considering a local neighborhood $ \mathbf {N} (i, j) $ around $(i,j)$ in the difference image $\mathbf{I}_d$. Mathematically, we can defined this process as:
\begin{equation}
    \mathbf{P}_c  = f_{\theta_{\text{D-CPG}}} (\mathbf{I}_d),
\end{equation}
where $f_{\theta_{\text{D-CPG}}} (\cdot)$ is the parametric representation of D-CPG. The size of the local-neighborhood $ \mathbf {N} (i, j) $ is usually defined by the size of the receptive field \cite{luo2016understanding} of D-CPG, and can be appropriately controlled by changing the network depth. We conduct experiments with different ConvNet architectures for our D-CPG like hourglass network \cite{newell2016stacked}, U-Net\cite{ronneberger2015u} with skip connections, and Residual Networks (ResNet) \cite{he2016deep} with different depth (i.e., 4, 8, 16, 32, and 64 residual blocks). The parameters of the D-CPG are initialized randomly and optimized over each iteration for a given MT-RSI during testing (see Fig. \ref{fig:itterative_opt}) by minimizing the proposed unsupervised loss $\mathcal{L}_{\text{CD}}$ function which will be introduced in the following section.

\subsection{Proposed unsupervised loss function for CD}
The proposed unsupervised loss function for CD $\mathcal{L}_{\text{CD}}$ consists of three terms, namely image-domain and feature-domain similarity-dissimilarity loss ($\mathcal{L}_{\text{img}}$ and $\mathcal{L}_{\text{feat}}$) and context consistency loss ($\mathcal{L}_{\text{ctx}}$) as detailed below. 

\paragraph{Image-domain loss for CD: $\mathcal{L}_{\text{img}}$.}
\label{sec:img_loss}
We propose an unsupervised loss function for CD, motivated by the most popular principle in \textit{metric learning}: minimize the distance between similar datasets and maximize the distance between dissimilar datasets \cite{qian2018large,tang2011partially,xing2002distance}. We adopt this notion to build our unsupervised loss for CD where we try to maximize the distance between change pairwise pixels while minimizing the distance between no-change pairwise pixels. We weight $\widetilde{\mathbf{I}}_1$ and $\widetilde{\mathbf{I}}_2$ according to the change probability $\mathbf{P}_{\text{c}}$ and no change probability $\mathbf{P}_{\text{nc}}$ $(= 1 - \mathbf{P}_{\text{c}})$ to identify  change and no change pixels, and then use the Mahalanobis distance to calculate the change loss $\mathcal{L}_{\text{img-c}}(\widetilde{\mathbf{I}}_1, \widetilde{\mathbf{I}}_2, \mathbf{P}_{\text{c}})$ and no-change loss $\mathcal{L}_{\text{img-nc}}(\widetilde{\mathbf{I}}_1, \widetilde{\mathbf{I}}_2, \mathbf{P}_{\text{nc}})$ as follows:
\begin{equation}
\begin{aligned}
	\label{image_loss}
    \mathcal{L}_{\text{img}} &= \alpha_{\text{img}} \mathcal{L}_{\text{img-c}}(\widetilde{\mathbf{I}}_1, \widetilde{\mathbf{I}}_2, \mathbf{P}_{\text{c}}) +  \mathcal{L}_{\text{img-nc}}(\widetilde{\mathbf{I}}_1, \widetilde{\mathbf{I}}_2, \mathbf{P}_{\text{nc}})\\
    &= - \alpha_{\text{img}}  \left\|  \mathbf{P}_{\text{c}} (\widetilde{\mathbf{I}}_1 - \widetilde{\mathbf{I}}_2)\right\|_{\mathbf{M}} +  \left\|  \mathbf{P}_{\text{nc}} (\widetilde{\mathbf{I}}_1 - \widetilde{\mathbf{I}}_2)\right\|_{\mathbf{M}},
\end{aligned}
\end{equation}
where we refer $\mathcal{L}_{\text{img}}$ to as image-domain change-nochange loss for CD, and $\alpha_{\text{img}}$ is the regularization parameter for balancing the trade-off between image-domain change and no-change loss terms. By default we set $\alpha_{\text{img}}$ to 1.0. When $\mathcal{L}_{\text{img}}$ is minimized, the no change loss $\mathcal{L}_{\text{img-nc}}$ will lead to minimize the distance between no-change pairwise pixels, and change loss $\mathcal{L}_{\text{img-c}}$ will lead to  maximize the distance between change pairwise pixels. Since $\mathcal{L}_{\text{img}}$ is differentiable, it back-propagates through the D-CPG and adjusts its parameters to generate better change map until it  converges.

\paragraph{Feature-domain loss for CD: $\mathcal{L}_{\text{feat}}$.} To impose context-based perspective of change maps in optimization, we adopt transfer learning \cite{pan2009survey} capability for CD. Transfer learning enables us to extract deep-features from a multi-layer deep feature extractor (D-FE) that is pre-trained for some other task. Such deep feature representations can capture spatial context of an image effectively. Therefore, to improve context-based perspective of predicted change map, we extend our change-nochange loss function from image-domain to feature-domain by utilizing the multi-layer feature representations of multi-temporal image $\{\mathbf{I}_1, \mathbf{I}_2\}$ as follows:
\begin{equation}
\label{eq:feat_loss}
\scriptsize
\begin{aligned}
	\mathcal{L}_{\text{feat}} &= \sum_{l=1}^{L} \left( \alpha_{\text{f}}^{l} \mathcal{L}_{\text{feat-c}}(\mathbf{F}_1^{l}, \mathbf{F}_2^{l}, d^l(\mathbf{P}_{\text{c}})) +  \mathcal{L}_{\text{feat-nc}}(\mathbf{F}_1^{l}, \mathbf{F}_2^{l}, d^l(\mathbf{P}_{\text{nc}})) \right) \\
    &= \sum_{l=1}^{L} \left( -\alpha_{\text{feat}}^{l}  \left\|  d^l(\mathbf{P}_{\text{c}}) (\mathbf{F}_1^{l} - \mathbf{F}_2^{l})\right\|_{\mathbf{M}} +   \left\|  d^l(\mathbf{P}_{\text{nc}}) (\mathbf{F}_1^{l} - \mathbf{F}_2^{l})\right\|_{\mathbf{M}} \right),
\end{aligned}
\end{equation}
where $l$ denotes the $l$-th layer of $L$-layered D-FE, $\alpha_{\text{feat}}^l$ is the regularization parameter for the $l$-th layer that is set to 1.0, $\mathcal{L}_{\text{feat-c}}(\cdot)$ is the feature-domain change loss, $\mathcal{L}_{\text{feat-nc}}(\cdot)$ is the feature-domain no-change loss, $d^l(\cdot)$ is the nearest-neighbor down-sampling of $\{\mathbf{P}_c, \mathbf{P}_{nc}\}$ to the spatial size of $\mathbf{F}_1^l$ or $\mathbf{F}_2^l$, respectively. 

\paragraph{Context consistency loss for CD: $\mathcal{L}_{\text{ctx}}$.}
Although deep features can provide contextual information for an image to a great extent, they are still sensitive to different illumination conditions, different image sensors, and noise. To make our estimated change map $\mathbf{P}_c$ to be further robust to such unwanted changes, we apply random augmentations on the multi-temporal image $\{\mathbf{I}_1, \mathbf{I}_2\}$ at each iteration of testing, process them though the deep feature extractor, and then constrain these features to be consistent with their original feature representations. In particular, we apply random color augmentations such as random brightness, contrast, saturation and hue jittering, as well as random noise augmentation \cite{shorten2019survey}. We denote the augmented versions of pre-change and post-change images at $t$-th iteration of optimization as $[\hat{\mathbf{I}}_{1}]_t$ and $[\hat{\mathbf{I}}_{2}]_t$, and their respective features from the $l$-th layer of D-FE as $[\hat{\mathbf{F}}^l_{1}]_t$ and $[\hat{\mathbf{F}}^l_{2}]_t$, respectively. We then impose context consistency between the original feature representations and the augmented feature representations by context consistency loss $\mathcal{L}_{\text{ctx}}$ as follows:
\begin{equation}
    \mathcal{L}_{\text{ctx}} = \sum_{l=1}^{L} \left( \left\| \mathbf{F}_1^{l} - [\hat{\mathbf{F}}^l_{1}]_t  \right\|_1 + \left\| \mathbf{F}_2^{l} - [\hat{\mathbf{F}}^l_{2}]_t  \right\|_1 \right), 
    \label{eq:ctx_loss}
\end{equation}
where $\left\| \cdot \right\|_1$ denotes the $L_1$ norm.

\paragraph{Sparsity penalty: $\mathcal{L}_{sparse}$.} A trivial solution exists for the objective (\ref{eq:feat_loss}) where, $\mathbf{P}_c = \mathbf{1}$. To avoid such degenerate solutions, we introduce a sparsity penalty $\mathcal{L}_{sparse}$ in the form of $1/\sin(\cdot)$ that discourages D-CPG  predicting all-one (or all zero); specifically,
\begin{equation}
    \mathcal{L}_{sparse} = \sin \left( \frac{\pi}{hw} \sum_{i=1}^{h} \sum_{j=1}^{w} P_c(i,j)\right)^{-1},
\end{equation}
where $(h,w)$ is the (height, width) of the change probability map $\mathbf{P}_c$.

The overall unsupervised loss function for CD $\mathcal{L}_{\text{CD}}$ that we use to optimize D-CPG and D-FE is defined as follows:
\begin{equation}
    \label{overall_loss}
    \mathcal{L}_{\text{CD}} = \mathcal{L}_{\text{img}} +  \mathcal{L}_{\text{feat}} + \mathcal{L}_{\text{ctx}} + \mathcal{L}_{sparse}.
\end{equation}

By minimizing $\mathcal{L}_{\text{CD}}$ using an optimizer such as Adam~\cite{kingma2014adam} over the parameters of D-CPG $\left( \theta_{\text{D-CPG}} \right)$ and D-FE $\left( \theta_{\text{D-FE}} \right)$, we obtain optimum change probability map $\mathbf{P}^*_c$ for a given bi-temporal image $\{\mathbf{I}_1, \mathbf{I}_2\}$, starting from the difference image $\mathbf{I}_d$. Hence, the empirical information available to the change detection process is the bi-temporal image $\{\mathbf{I}_1, \mathbf{I_2}\}$ and pre-trained weights of D-FE $\theta_{\text{DFE}}^0$ from transfer learning, and hence the CD process is completely unsupervised. This approach is schematically depicted in Fig. \ref{fig:itterative_opt} and Fig. \ref{fig:method}.

\begin{table*}[tb]
    \caption{The average quantitative CD results on the OSCD dataset~\cite{OSCD_dataset}, the SZTAKI AirChange dataset~\cite{airc_dataset1,airc_dataset2}, and the QuickBird dataset~\cite{QuickBird}. Higher values of OA, UA, Recall, F1, and AUC indicate good change detection performance. Color convention: \textcolor{red}{\bf Best}, \textcolor{blue}{\bf 2-nd Best}, \textcolor{black}{\bf 3-rd Best}.}
    \label{tab:quantitative_results}
    \centering
    \resizebox{\textwidth}{!}{%
	\begin{tabular}{p{2.3cm}p{0.9cm}p{0.9cm}p{0.9cm}p{0.9cm}p{0.9cm}cp{0.9cm}p{0.9cm}p{0.9cm}p{0.9cm}p{0.9cm}cp{0.9cm}p{0.9cm}p{0.9cm}p{0.9cm}p{0.9cm}}
		\toprule
		\multirow{2}{*}{\parbox[c]{.2\linewidth}{\centering Method}} & \multicolumn{5}{c}{OSCD Dataset~\cite{OSCD_dataset}} & & \multicolumn{5}{c}{SZTAKI AirChange Dataset~\cite{airc_dataset1, airc_dataset2}} & & \multicolumn{5}{c}{QuickBird Dataset~\cite{QuickBird}} \\ 
		\cmidrule{2-6} \cmidrule{8-12} \cmidrule{14-18}
		            &{\textbf{OA}}&{\textbf{UA}}&{\textbf{Recall}}&{\textbf{F1}}&{\textbf{AUC}}&&     {\textbf{OA}}&{\textbf{UA}}&{\textbf{Recall}}&{\textbf{F1}}&{\textbf{AUC}}&&     {\textbf{OA}}&{\textbf{UA}}&{\textbf{Recall}}&{\textbf{F1}}&{\textbf{AUC}}\\
		\midrule
	    \multicolumn{8}{l}{\bf Unsupervised CD methods:}
		\\
		CVA~\cite{cva} 	            &0.830  & 0.164  & 0.457 & 0.192 & 0.751 &&                
		                            0.768 & 0.118 & 0.545 & 0.179 & 0.708 && 
		                            0.858 & 0.415 & 0.699 & 0.521 & 0.865\\
		DPCA~\cite{dpca} 	        & \bf 0.929 & 0.238 & 0.238 & 0.189 & 0.823 &&                
		                            0.861 & 0.167 & 0.377 & 0.191 & 0.704&& 
		                            \bf 0.893 & \textcolor{blue}{\bf 0.512} & 0.508 & 0.510 & 0.807\\
		ImageDiff~\cite{ImageDiff}  & 0.799 & 0.148 & \textcolor{blue}{\bf 0.483} & 0.181 & 0.752    &&                
		                            0.769 & 0.121 & 0.546 & 0.182 & 0.696 &&
		                            0.857 & 0.411 & 0.695 & 0.516 &  0.861\\
		ImageRatio~\cite{ImageRatio}& 0.810 & 0.113 & 0.402 & 0.149 & 0.697&&            
		                            0.863 & 0.067 & 0.136 & 0.042 & 0.604 && 
		                            0.829 & 0.352 & 0.667 & 0.461 &  0.807\\
		ImageRegr~\cite{ImageRegr} 	& 0.897 & 0.196 & 0.413 & \bf 0.222 & 0.821   &&               
		                            0.747 & 0.134 & \textcolor{red}{\bf 0.615} & 0.198 & \bf 0.743 &&  
		                            0.870 & 0.437 & 0.647 & 0.522 & 0.849\\
		IRMAD~\cite{IRMAD} 		    & 0.900 & 0.155 & 0.472 & 0.204 & \textcolor{blue}{\bf 0.864}  &&                
		                            0.822 & 0.162 & 0.507 & \bf 0.223 &  0.722 && 
		                            0.843 & 0.394 & \textcolor{red}{\bf 0.789} & 0.526 & \bf 0.885\\
		MAD~\cite{MAD} 		        & 0.877 & 0.143 & \textcolor{red}{\bf 0.534} & 0.191 & \textcolor{black}{\bf 0.857}&&                
		                            0.770 & 0.121 & \bf 0.594 & 0.189 & 0.714&& 
		                            0.815 & 0.348 & \textcolor{blue}{\bf 0.782} & 0.482 & 0.871\\
		PCDA~\cite{PCDA} 		    & 0.855 & 0.192 & 0.456 & 0.210 & 0.805&&                
		                            0.758 & 0.119 & \textcolor{blue}{\bf 0.598} & 0.183 & 0.737&& 
		                            0.862 & 0.425 & \bf 0.728 & \bf 0.537 & 0.889\\
		DeepCVA~\cite{DeepCVA} 		& 0.916 & 0.216 & 0.350 & \textcolor{blue}{\bf 0.245} & 0.835&&                
		                            0.859 & \textcolor{blue}{\bf 0.259} & 0.277 & \textcolor{blue}{\bf 0.268} & \textcolor{blue}{\bf 0.847}&& 
		                            0.863 & 0.432 & 0.544 & 0.482 & \textcolor{blue}{\bf 0.891}\\
		\hline
	    \multicolumn{18}{l}{\textbf{Supervised CD methods} (supervised on the OSCD training set and tested on the test set of the respective dataset):}
		\\
		UNet~\cite{unet}            & \textcolor{blue}{\bf 0.946} & \textcolor{blue}{\bf 0.466} & 0.176 & 0.186 & 0.776 &&                
		                            \textcolor{blue}{\bf 0.936} & 0.187 & 0.098 & 0.088 & 0.718&& 
		                            0.844 & 0.344 & 0.355 & 0.349 & 0.839\\
		SeCo-(Rand)~\cite{SeCo} 		& \textcolor{blue}{\bf 0.946} & \textcolor{red}{\bf 0.477} & 0.150 & 0.180 & 0.812 &&                
		                            \bf 0.907 & \bf 0.234 & 0.306 & 0.210 & 0.720&& 
		                            0.880 & \bf 0.506 & 0.488 & 0.497 &  0.777\\
		SeCo-(Pre)~\cite{SeCo} 		& 0.928 & \bf 0.412 & 0.182 & 0.195 & 0.747&&
		                            0.860 & 0.198 & 0.363 & 0.199 & 0.691&&
		                            \textcolor{blue}{\bf 0.905} & \textcolor{red}{\bf 0.654} & 0.588 & \textcolor{red}{\bf 0.619} & 0.867\\
		\hline
		\multicolumn{18}{l}{\textbf{Deep Metric Learning CD}:}
		\\
		\bf Ours 		        & \textcolor{red}{\bf 0.958} & 0.383 & \textcolor{black}{\bf 0.480} & \textcolor{red}{\bf 0.325} & \textcolor{red}{\bf 0.937}&&                
		                            \textcolor{red}{\bf 0.948} & \textcolor{red}{\bf 0.325} & 0.256 & \textcolor{red}{\bf 0.286} & \textcolor{red}{\bf 0.913}&& 
		                            \textcolor{red}{\bf 0.908} & 0.486 & 0.697 & \textcolor{blue}{\bf 0.573} & \textcolor{red}{\bf 0.928}\\
		\bottomrule
	\end{tabular}}
\end{table*}

\section{Experimental Setup}
\paragraph{Datasets.} We evaluate the performance of our proposed CD framework on three widely used and publically available CD datasets, namely Onera Satellite Change Detection Dataset (abbreviated as OCSD) ~\cite{OSCD_dataset}, SZTAKI AirChange Benchmark set (abbreviated as SZTAKI)~\cite{airc_dataset1, airc_dataset2}, and QuickBird dataset~\cite{QuickBird}. The \textbf{OSCD dataset} consists of 24 pairs of 13-band ($b$ = 13) multispectral images taken from the Sentinel-2 satellites~\cite{sentinel-2}, captured all over the world (such as Brazil, the US, Europe, the Middle East, and Asia)  between 2015 and 2018.  Note that we present results on the testing set of the OSCD dataset. In addition, the \textbf{SZTAKI dataset} contains 13 aerial image pairs of size $952 \times 640$ and the  corresponding binary change masks. The bi-temporal images are taken with a 23-years of time difference. We use all 13 image pairs of the SZTAKI dataset for testing. Furthermore, the \textbf{QuickBird dataset} consists of a three-band bi-temporal image pair of size $1154 \times 740$ which was captured in 2009 and 2014. In the QuickBird dataset, seasonal changes are very prominent - making it difficult to filter out relevant changes from unwanted ones compared to the other two datasets.

\paragraph{Implementation details and reproducibility.}
We implemented our unsupervised CD framework in Pytorch with an NVIDIA Quadro RTX 8000 GPU. As we discussed in Sec. \ref{sec: method}, our CD framework does not involve a training phase, unlike supervised or self-supervised CD approaches. Instead, we iteratively optimize the parameters of the network for each MT-RSI separately. After a fixed number of iterations (set to 80), we save the resulting change probability map from the network (note that we do not save the network parameters), and obtain a binary change map by thresholding it with an appropriate probability (set to 0.5, but can manually vary for better performance). We initialize multi-level D-FE from VGG-16~\cite{VGG} pre-trained weights on ImageNet~\cite{ImageNet}. Additional experiments on how CD performance varies with different initialization methods can be found in the supplementary document. For all the experiments, we used Adam optimizer with a learning rate of 1e-5. 
\textit{Implementation code will be made publicly available after the review process.}

\paragraph{Performance metrics and visualization of change maps.} We adopt the overall accuracy (\textbf{OA}), user accuracy (\textbf{UA}) - often called as precision, recall rate (\textbf{Recall}), F-1 measure (\textbf{F1}), and Area Under the Receiver Operating Characteristic (ROC) curve (\textbf{AUC}) to \textit{quantitatively} compare the CD performance of different methods. Among these, AUC is the best way to summarize a  model's ability to distinguish between change and no change regions. A higher value of AUC indicates a good CD model. For the \textit{qualitative} comparisons, we highlight each pixel in the predicted change map with different colors to denote true positives (\textcolor{ForestGreen}{\textbf{TP}}), true negatives (\colorbox{Black}{\textcolor{White}{\textbf{TN}}}), false positives (\textcolor{BrickRed}{\textbf{FP}}), and false negatives (\textcolor{Blue}{\textbf{FN}}). Thus, a change map with \textit{more} \textcolor{Green}{\textbf{green}} and \colorbox{Black}{\textcolor{White}{\textbf{white}}} pixels, and \textit{fewer} \textcolor{BrickRed}{\textbf{red}} and \textcolor{Blue}{\textbf{blue}} pixels indicates a good CD performance.

\begin{figure*}[tbh]
    \centering
    \begin{minipage}{0.32\textwidth}%
        \includegraphics[width=\linewidth]{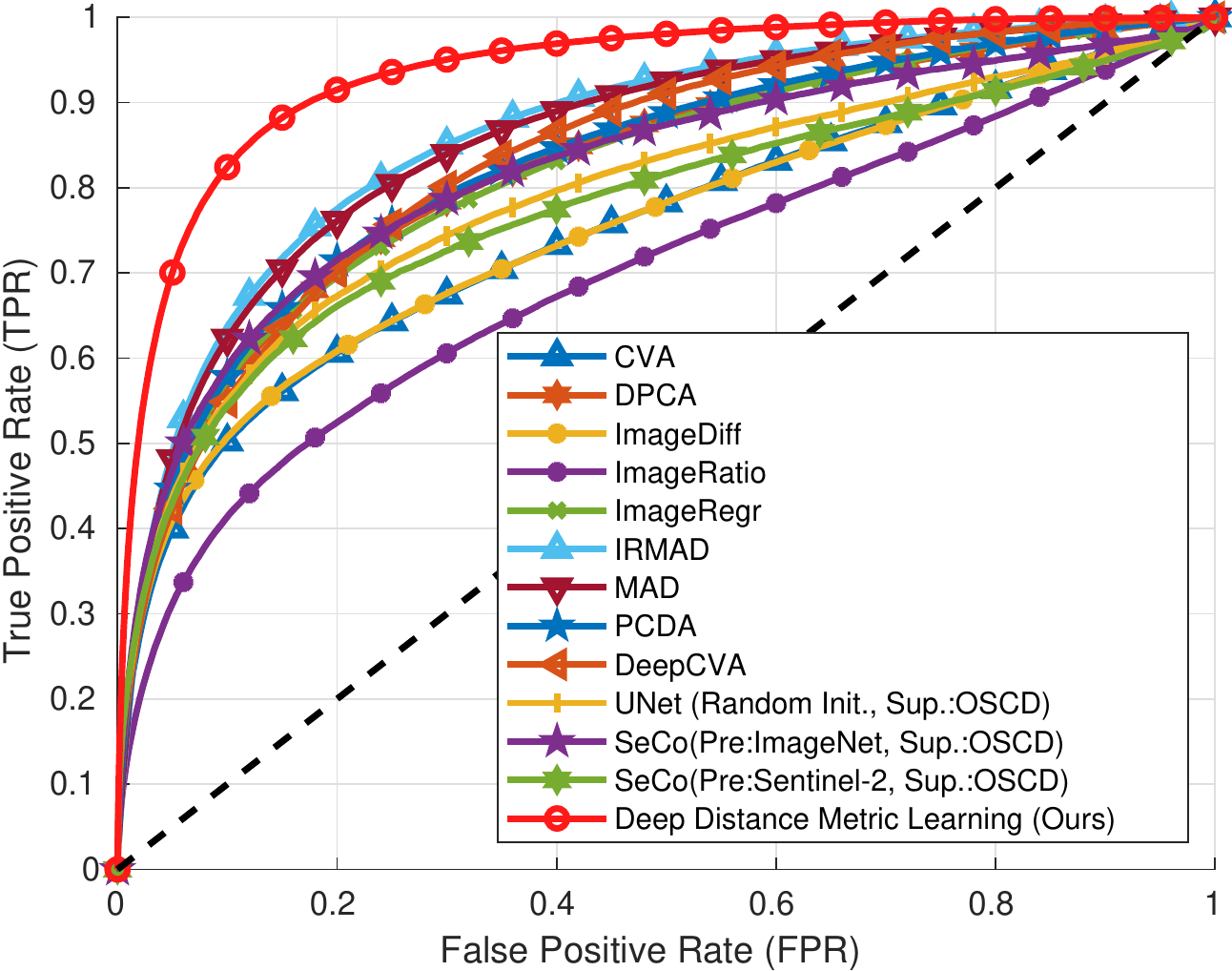}
        \label{fig:subfig1}
    \end{minipage}%
    \begin{minipage}{0.32\textwidth}%
        \includegraphics[width=\linewidth]{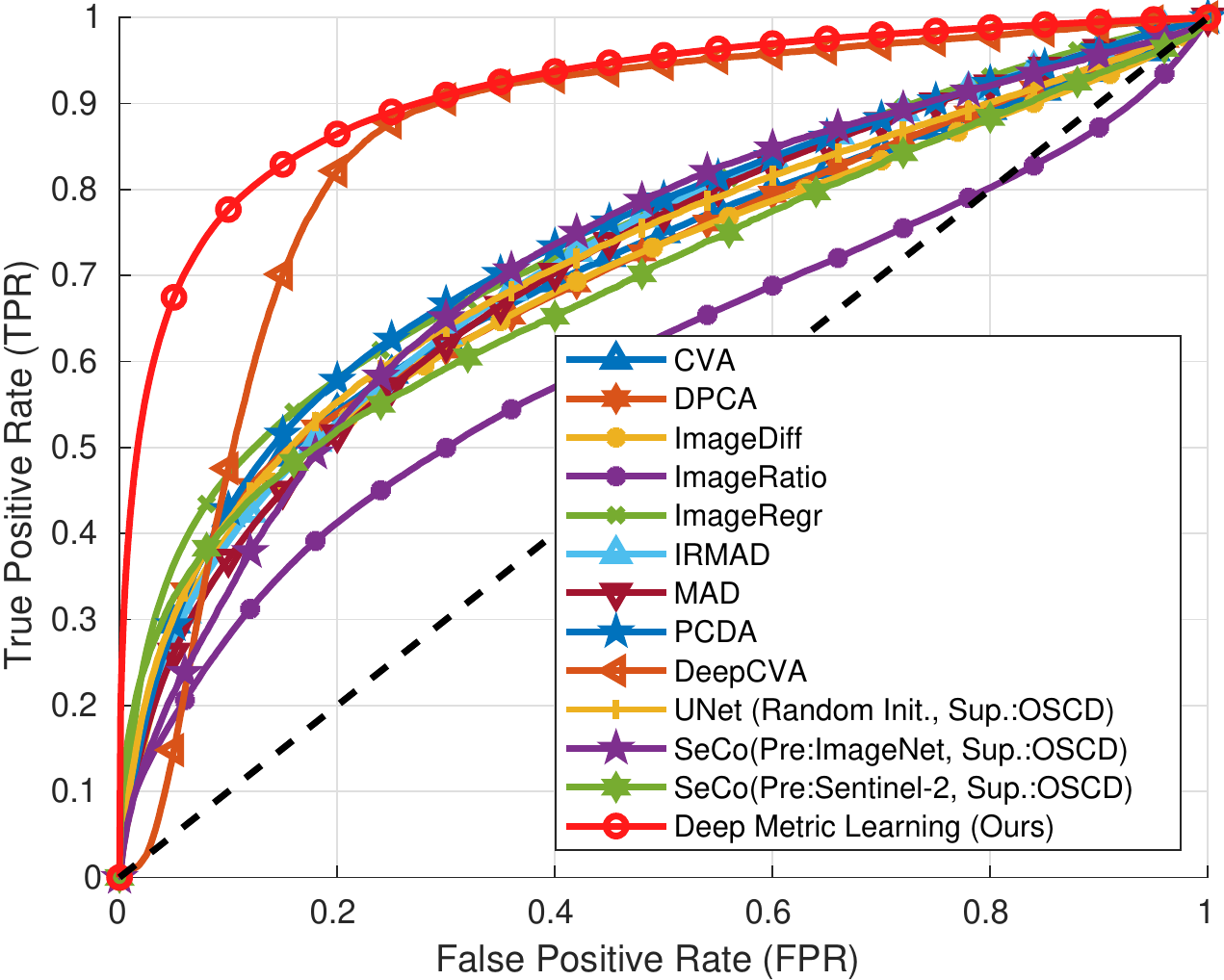}
        \label{fig:subfig2}
    \end{minipage}
    \begin{minipage}{0.32\textwidth}%
        \includegraphics[width=\linewidth]{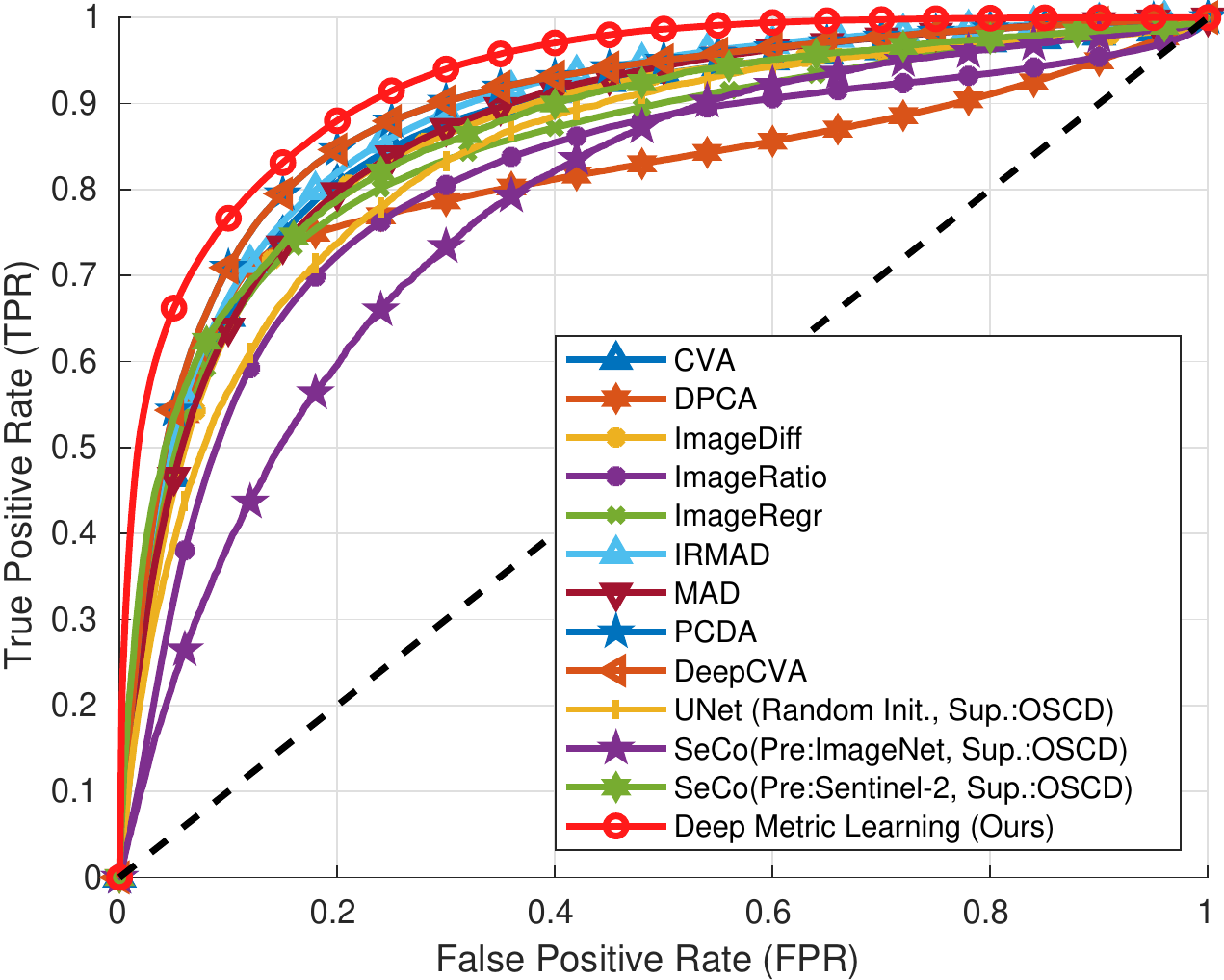}
        \label{fig:subfig3}
    \end{minipage}
    \caption{Average ROC curves corresponding to different CD methods on \textbf{(a)} OSCD~\cite{OSCD_dataset}, \textbf{(b)} SZTAKI~\cite{airc_dataset1, airc_dataset2}, and \textbf{(c)} QuickBird datasets~\cite{QuickBird}.}
    \label{fig:roc}
\end{figure*}

\begin{figure*}[tbh]
    \centering
    \includegraphics[width=\linewidth]{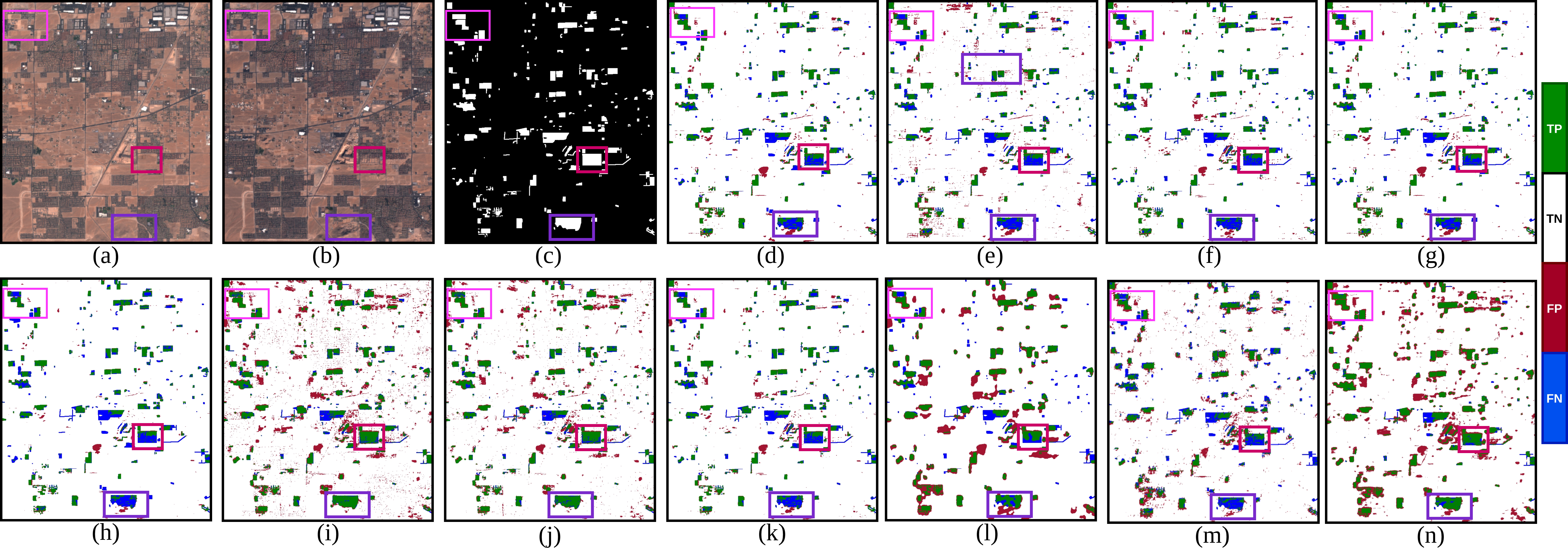}
    \caption{Qualitative results on {\ttfamily lasvegas} image in the OSCD dataset. \textbf{(a)} $\mathbf{I}_1$. \textbf{(b)} $\mathbf{I}_2$. \textbf{(c)} Reference change mask. \textbf{(d)} ImageDiff~\cite{ImageDiff}. \textbf{(e)} ImageRatio~\cite{ImageRatio}. \textbf{(f)} ImageRegr~\cite{ImageRegr}. \textbf{(g)} CVA~\cite{cva}. \textbf{(h)} DPCA~\cite{dpca}. \textbf{(i)} MAD~\cite{MAD}. \textbf{(j)} IRMAD~\cite{IRMAD}. \textbf{(k)} PCDA~\cite{PCDA}. \textbf{(l)} Deep-CVA~\cite{DeepCVA}. \textbf{(m)} SeCo~\cite{SeCo}: Self-supervised pre-trained on Sentinel-2~\cite{sentinel-2} and supervised on the OSCD~\cite{OSCD_dataset} training set. \textbf{(n)} Deep Metric Learning CD (Ours).
    }
    \label{fig:OSCD_cm}
\end{figure*}
\begin{figure*}[h]
    \centering
    \includegraphics[width=\linewidth]{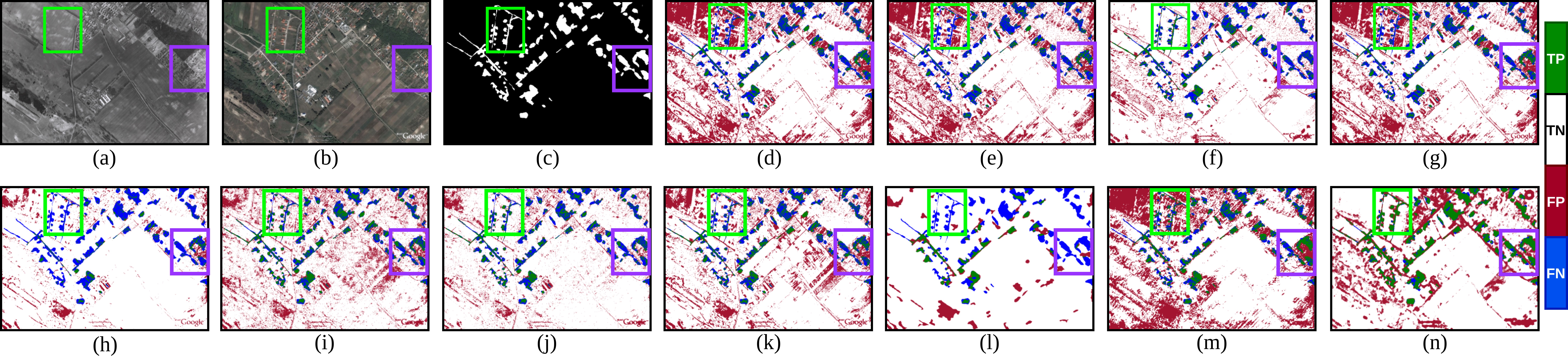}
    \caption{Qualitative results on the  {\ttfamily archive} image in the SZTAKI dataset~\cite{airc_dataset1, airc_dataset2}. \textbf{(a)} $\mathbf{I}_1$. \textbf{(b)} $\mathbf{I}_2$. \textbf{(c)} Reference change mask. \textbf{(d)} ImageDiff~\cite{ImageDiff}. \textbf{(e)} ImageRatio~\cite{ImageRatio}. \textbf{(f)} ImageRegr~\cite{ImageRegr}. \textbf{(g)} CVA~\cite{cva}. \textbf{(h)} DPCA~\cite{dpca}. \textbf{(i)} MAD~\cite{MAD}. \textbf{(j)} IRMAD~\cite{IRMAD}. \textbf{(k)} PCDA~\cite{PCDA}. \textbf{(l)} Deep-CVA~\cite{DeepCVA}. \textbf{(m)} SeCo~\cite{SeCo}: Self-supervised pre-trained on Sentinel-2~\cite{sentinel-2} and supervised on the OSCD~\cite{OSCD_dataset} training set. \textbf{(n)} Deep Metric Learning CD (Ours).}
    \label{fig:SZTAKI_cm}
\end{figure*}
\begin{figure*}[tbh]
    \centering
    \includegraphics[width=\linewidth]{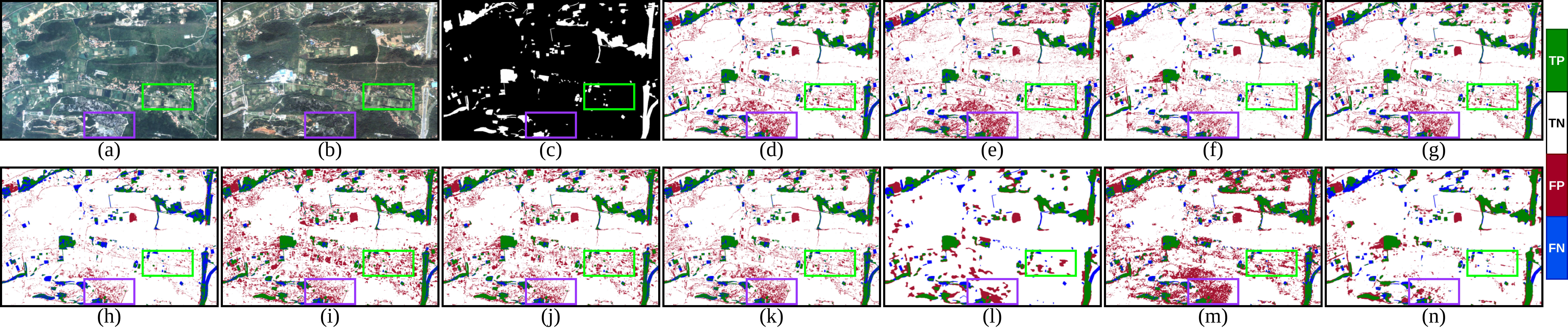}
    \caption{Qualitative results on the QuickBird dataset~\cite{QuickBird}. \textbf{(a)} $\mathbf{I}_1$. \textbf{(b)} $\mathbf{I}_2$. \textbf{(c)} Reference change mask. \textbf{(d)} ImageDiff~\cite{ImageDiff}. \textbf{(e)} ImageRatio~\cite{ImageRatio}. \textbf{(f)} ImageRegr~\cite{ImageRegr}. \textbf{(g)} CVA~\cite{cva}. \textbf{(h)} DPCA~\cite{dpca}. \textbf{(i)} MAD~\cite{MAD}. \textbf{(j)} IRMAD~\cite{IRMAD}. \textbf{(k)} PCDA~\cite{PCDA}. \textbf{(l)} Deep-CVA~\cite{DeepCVA}. \textbf{(m)} SeCo~\cite{SeCo}: Self-supervised pre-trained on Sentinel-2~\cite{sentinel-2} and supervised on the OSCD~\cite{OSCD_dataset} training set. \textbf{(n)} Deep Metric Learning CD (Ours).
    }
    \label{fig:QuickBird_cm}
\end{figure*}

\section{Results and Discussion}
\paragraph{Quantitative results.}
Table \ref{tab:quantitative_results} presents average quantitative results corresponding to different CD methods. From these results, we can make the following key observations: \textbf{1.} The proposed CD method outperforms the  existing unsupervised approaches (classical and deep) by a significant margin, especially in OA, F1 and AUC metrics. \textbf {2.} It even outperforms recent deep supervised approaches, which are pre-trained in an unsupervised manner on a large RS (SeCo- (P)~\cite{SeCo}) dataset and then fine-tuned on the OSCD training set in a supervised manner. The latter observation proves the point we made in the introduction, where the supervised CD approaches are weak at generalizing under domain shifts, which is common in the datasets we consider here. Since the proposed CD approach addresses this problem by optimizing the network parameters for a given MT-RSI separately to obtain the change probability map, it shows consistently higher CD performance in OA, F1, and AUC metrics than the SOTA approaches. Furthermore, we visualize the ROC curves corresponding to different CD methods in Fig. \ref{fig:roc}. From these graphs, one can clearly see the significance of the proposed CD approach compared to SOTA CD methods.

\paragraph{Qualitative results.} We present qualitative visualizations of the predicted change maps from different CD algorithms on the OSCD, SZTAKI, and QuickBird datasets in Fig. \ref{fig:OSCD_cm}, Fig. \ref{fig:SZTAKI_cm}, and Fig. \ref{fig:QuickBird_cm}, respectively. We can see that the proposed CD can capture most of the relevant changes that are difficult to capture by the SOTA CD methods (see the highlighted regions by bounding boxes) while minimizing false positives and false negative predictions. Please see supplementary document for additional qualitative results. 

\begin{figure}[!htb]
    \centering
    \includegraphics[width=\linewidth]{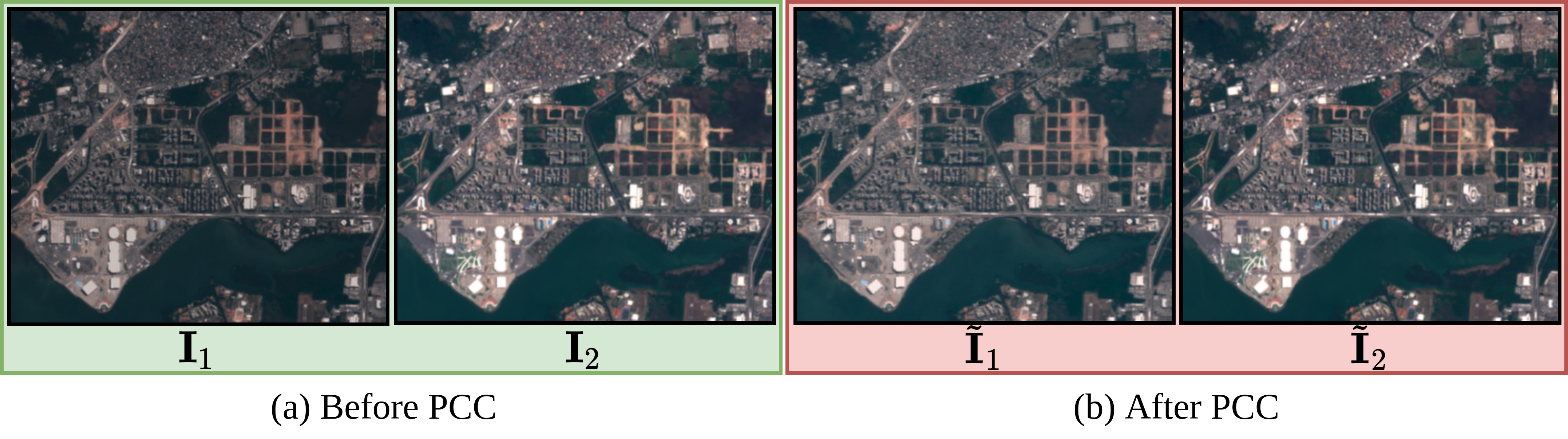}
    \caption{
    The effect of PCC on {\ttfamily rio} image pair in the OSCD~\cite{OSCD_dataset}.
    }
    \label{fig:ablation_pcc}
\end{figure}

\section{Ablation Study}
\paragraph{Effect of PCC.} Fig. \ref{fig:ablation_pcc} shows an example of how PCC minimizes the global and local color changes in MT-RSIs. After PCC, the pre-change image's $(\mathbf{I}_1)$ color space is mapped to the post-change image's $(\mathbf{I}_2)$ color space - making the difference image $(\mathbf{I}_d)$ to be less sensitive to colorimetric changes. Please see supplementary document for additional examples.

\begin{figure}[tbh!]
    \centering
    \includegraphics[width=\linewidth]{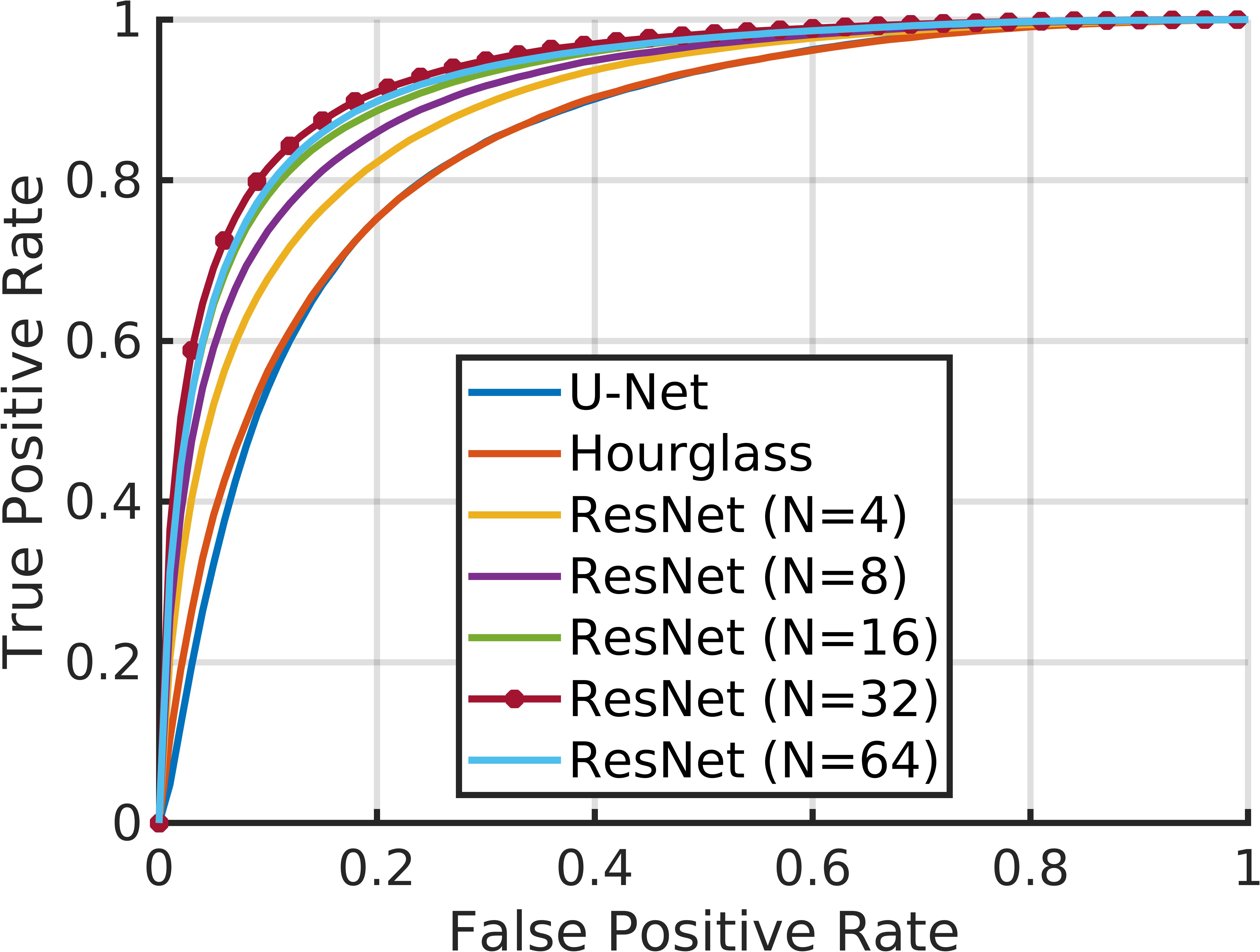}
    \caption{Average ROC curves for different architectures of D-CPG on OSCD~\cite{OSCD_dataset}.}
    \label{fig:ab_dcpg}
\end{figure}

\begin{table}[tbh!]
    \centering
    \begin{tabular}{lc|l}
        \toprule
        Network         && \textbf{AUC}   \\ 
        \midrule
        U-Net           && 0.847 \\
        Hourglass       && 0.852 \\
        ResNet (N=4)    && 0.890 \\
        ResNet (N=8)    && 0.908 \\
        ResNet (N=16)   && \textcolor{black}{\bf 0.922} \\
        ResNet (N=32)   && \textcolor{red}{\bf 0.937} \\
        ResNet (N=64)   && \textcolor{blue}{\bf 0.925} \\ 
        \bottomrule
    \end{tabular}
    \caption{Caption}
    \label{ab:d_cpg_auc}
\end{table}

\paragraph{Experiments with different architectures for D-CPG.} We experimented with different network architectures for D-CPG, and the results are summarized in Fig. \ref{fig:ab_dcpg} and Tab. \ref{ab:d_cpg_auc}. We observe a higher AUC when we cascade 32 residual blocks (denoted as ResNet ($N=32$)) compared to Hourglass, U-Net, and other variations of ResNet. Furthermore, we generally observe faster convergence when we utilize ResNet architecture than U-Net and Hourglass networks. Therefore, for all of our experiments, we utilized ResNet with $N=32$.

\begin{figure}[!tbh]
    \centering
    \includegraphics[width=\linewidth]{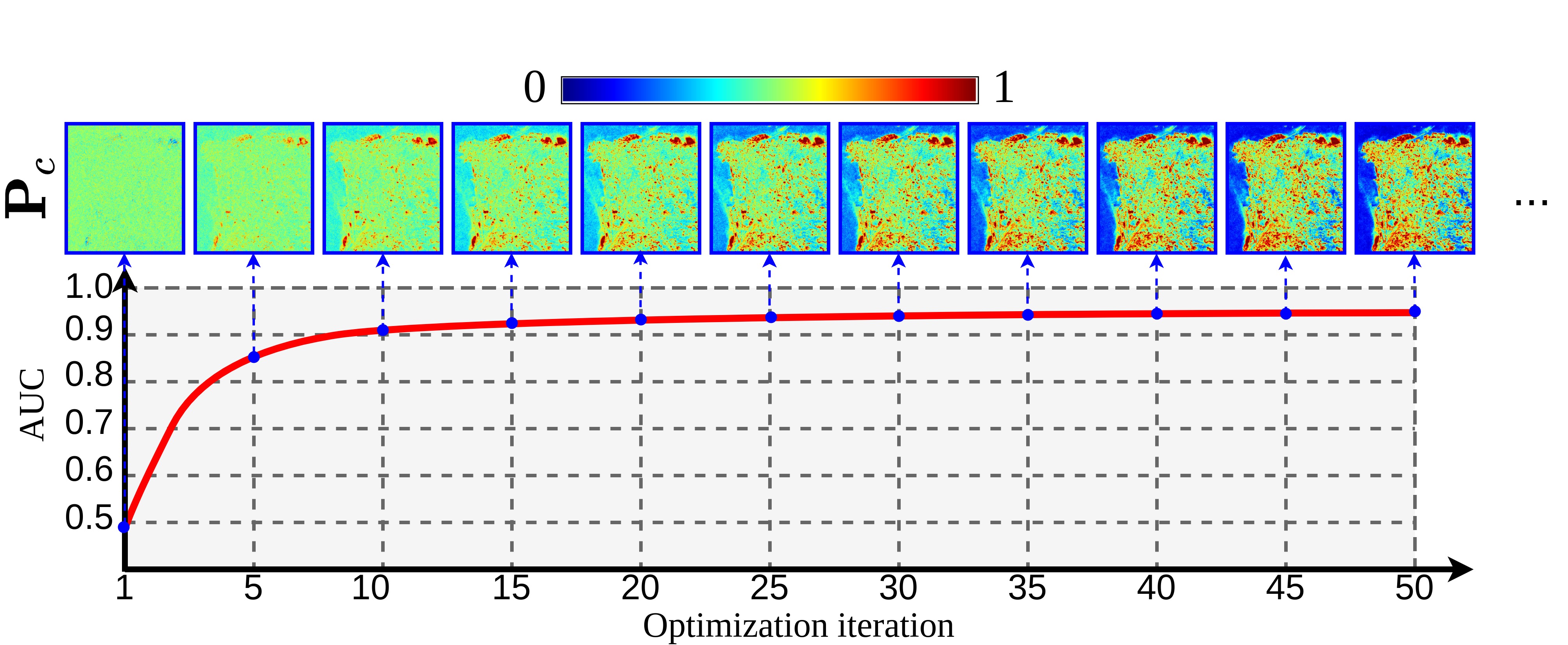}
    \caption{How our CD method converges to the optimal change map $\mathbf{P}^*_{\text{c}}$ during testing on the {\ttfamily beirut} image pair in OSCD~\cite{OSCD_dataset}.
    }
    \label{fig:convergence}
\end{figure}

\paragraph{Convergence characteristic.} Fig. \ref{fig:convergence} shows an example of how the proposed approach converges to the optimal change probability map $\mathbf{P}^*_c$ for a given bi-temporal image pair.

\begin{table}[tb]
    \centering
    \caption{Contribution from each loss term in $\mathcal{L}_{\text{CD}}$ on the CD performance for $l \in \{1,2\}$(averaging over the test-set of OSCD~\cite{OSCD_dataset}).}
    \begin{tabular}{ccc|ccc}
    \toprule
    $\mathcal{L}_{\text{img}}$ &  $\mathcal{L}_{\text{feat}}$ & $\mathcal{L}_{\text{ctx}}$ & \textbf{OA} & \textbf{F1} & \textbf{AUC}\\
    \midrule
    \cmark      & \xmark    &  \xmark   & 0.911 & 0.222 & 0.821\\
    \xmark      & \cmark    &  \xmark   & 0.936 & 0.296 & 0.879\\
    \xmark      & \cmark    & \cmark    & \textcolor{blue}{\bf 0.953} & \textcolor{blue}{\bf 0.319} & \textcolor{blue}{\bf 0.926}\\
    \cmark      & \cmark    &  \xmark   & \textcolor{black}{\bf 0.940} & \textcolor{black}{\bf 0.303} & \textcolor{black}{\bf 0.892}\\
    \cmark      & \cmark    & \cmark    & \textcolor{red}{\bf 0.958} & \textcolor{red}{\bf 0.325} & \textcolor{red}{\bf 0.937}\\
    \bottomrule
    \end{tabular}
    \label{tab:ab_loss}
\end{table}

\paragraph{Contribution from different loss terms.} Table \ref{tab:ab_loss} summarizes the contribution from each loss term - $\mathcal{L}_{\text{img}}$, $\mathcal{L}_{\text{feat}}$, and $\mathcal{L}_{\text{ctx}}$ on the CD performance. We can observe that computing similarity-dissimilarity loss in the feature domain ($\mathcal{L}_{\text{feat}}$) improves the CD performance in OA, F1, and AUC metrics significantly over the image domain loss ($\mathcal{L}_{\text{img}}$). Furthermore, imposing context-consistency constraint on deep features ($\mathcal{L}_{\text{ctx}}$) on top of $\mathcal{L}_{\text{feat}}$ results in a significant boost in the CD performance. Moreover, combining the three loss terms - $\mathcal{L}_{\text{img}}$, $\mathcal{L}_{\text{feat}}$, and $\mathcal{L}_{\text{ctx}}$ with appropriate regularization constants further improves the CD performance.

\begin{table}[t]
    \centering
    \caption{Effect of utilizing different scales of deep features from D-FE on the CD performance (averaging over test-set of OSCD~\cite{OSCD_dataset}).}
    \begin{tabular}{l|ccc}
    \toprule
    Loss function configuration & \textbf{OA} & \textbf{F1} & \textbf{AUC}\\
    \midrule
    Baseline: $\mathcal{L}_{\text{img}}$   & 0.911 & 0.222 & 0.821\\
    $\mathcal{L}_{\text{img}}$+$\mathcal{L}_{\text{feat}}^l+\mathcal{L}_{\text{ctx}}^{l}$; $l\in\{1\}$   & \bf 0.936 & 0.298 & 0.855\\
    $\mathcal{L}_{\text{img}}$+$\mathcal{L}_{\text{feat}}^l+\mathcal{L}_{\text{ctx}}^{l}$; $l\in\{1,2\}$   & \textcolor{red}{\bf 0.958} & \textcolor{red}{\bf 0.325} & \textcolor{red}{\bf 0.937}\\
    $\mathcal{L}_{\text{img}}$+$\mathcal{L}_{\text{feat}}^l+\mathcal{L}_{\text{ctx}}^{l}$; $l\in\{1,2,3\}$   & \textcolor{blue}{\bf 0.943} & \textcolor{blue}{\bf 0.315} & \textcolor{blue}{\bf 0.929}\\
    $\mathcal{L}_{\text{img}}$+$\mathcal{L}_{\text{feat}}^l+\mathcal{L}_{\text{ctx}}^{l}$; $l\in\{1,2,3,4\}$ & 0.932 & \bf 0.307 & \bf 0.914\\
    \bottomrule
    \end{tabular}
    \label{tab:ab_feat_scales}
\end{table}

\paragraph{Performance with different scales of features from D-FE.} According to the results summarized in Tab. \ref{tab:ab_feat_scales}, the best CD performance is observed when we utilize deep features from the first two hierarchical scales of VGG-16 (i.e., $l$=1 and 2). Utilizing additional scales of deep features result in degradation of the CD performance - so for all of our experiments we set $l=2$ in $\mathcal{L}_{\text{feat}}$ (Eqn. \ref{eq:feat_loss}) and $\mathcal{L}_{\text{ctx}}$ (Eqn. \ref{eq:ctx_loss}).


\section{Conclusion}
Society is becoming increasingly aware of human activities on the global climate. There is overwhelming evidence that these activities have short- and long-term effects on almost every aspect of our lives. Using global climate measurements and simulations, it is now possible to observe changes on a global scale, such as sea level rise or changes in the Gulf Stream. On the other hand, accurate predictions of local changes are much more difficult to obtain. Common examples include land use for agriculture, deforestation, flooding, forest fires, growth of urban areas, and transportation infrastructure. Therefore, it is extremely important to monitor these local changes as these are the factors that ultimately exacerbate the global climate crisis.

\bibliographystyle{splncs04}
\bibliography{egbib}

\end{document}